\documentclass[letterpaper, 10 pt, conference]{ieeeconf}  

\IEEEoverridecommandlockouts                              

\title{\LARGE \bf
Meta-Reinforcement Learning for Adaptive Motor Control in Changing Robot Dynamics and Environments
}

\author{Timothée Anne$^{1}$, Jack Wilkinson$^{2}$ and Zhibin Li$^{2}$
\thanks{$^{1}$Timothée Anne is with ENS Rennes, France; $^{2}$Jack Wilkinson and Zhibin Li are with the School of Informatics, the University of Edinburgh, UK. {\tt\small timothee.anne@ens-rennes.fr}, {\tt\small jack.wilkinson, zhibin.li@ed.ac.uk}}%
}
\usepackage{graphicx}
\graphicspath{ {fig/} }
\usepackage{amsmath}
\usepackage{amssymb}
\usepackage{algorithmic}
\usepackage{algorithm}
\usepackage{url}
\usepackage{subcaption}
\usepackage{graphicx}
\usepackage[dvipsnames]{xcolor}
\usepackage{balance} 
\usepackage{hyperref}
\hypersetup{
    colorlinks=true,
    linkcolor=blue,
    filecolor=magenta,      
    urlcolor=cyan,
}

\begin{document}
\bstctlcite{IEEEexample:BSTcontrol}

\maketitle
\thispagestyle{empty}
\pagestyle{empty}

\begin{abstract} 

This work developed a meta-learning approach that adapts the control policy on the fly to different changing conditions for robust locomotion. The proposed method constantly updates the interaction model, samples feasible sequences of actions of estimated the state-action trajectories, and then applies the optimal actions to maximize the reward. To achieve online model adaptation, our proposed method learns different latent vectors of each training condition, which are selected online given the newly collected data. Our work designs appropriate state space and reward functions, and optimizes feasible actions in an MPC fashion which are then sampled directly in the joint space considering constraints, hence requiring no prior design of specific walking gaits. We further demonstrate the robot's capability of detecting unexpected changes during interaction and adapting control policies quickly. The extensive validation on the SpotMicro robot in a physics simulation shows adaptive and robust locomotion skills under varying ground friction, external pushes, and different robot models including hardware faults and changes. 

\end{abstract}
\section{Introduction}
 

In robot motor control, developing responsive control policies that adapt to unforeseen environments is crucial to task success. These changes and unexpected situations can be intrinsic or extrinsic, such as robot damage, motor failure, varying friction, and external force disturbances. For robot locomotion, traditional approaches of planning and control require expert knowledge and accurate dynamics models and constraints of both the robot and the environment \cite{fankhauser2018robust, chatzinikolaidis2020}, which are all subject to unforeseeable changes that are difficult to know beforehand. Moreover, even using data-efficient learning techniques such as Bayesian optimization to tune decision variables and control parameters, it can only achieve adaptation on a trial-by-trial basis \cite{rai2018bayesian} and also require extensive computation \cite{yuan2019} which is not able to respond to changes on the fly.

Recent advances in Reinforcement-Learning (RL) lead to algorithms achieving human-like or animal-level performance in a range of difficult control tasks. Model-free RL can perform global search of control parameters and obtain globally optimal gaits while combined with walking pattern generation \cite{dallali2012global}. Also, an RL-based feedback policy can achieve human-like bipedal walking by imitating human motion capture data \cite{yang2020learning}. With simulation training including actuator properties, a model-free RL scheme can train different locomotion policies separately and deploy on a real quadrupedal robot \cite{hwangbo2019learning}. By using a multi-expert learning, an hierarchical RL architecture can learn to fuse multiple motor skills and generate multimodal locomotion coherently on a real quadruped \cite{yang2020multi}. However, in general, model-free RL algorithms have limited sample efficiency, resulting in long training time to produce viable policies. For example, it took a model-free RL algorithm 83 hours to achieve human performance on the Atari game suite, compared to 15 minutes for a human~\cite{hessel_rainbow_2017}. Similarly, AlphaStar~\cite{alphastarblog} used 200 years of equivalent real-time to reach expert human performance playing Starcraft II. 

On the other hand, model-based approaches can achieve comparatively high performance while being more sample-efficient by several orders of magnitude, converging faster than model-free approaches for locomotion tasks~\cite{Chua}. To deploy robots in the real-world, online adaptation to changes in the environment are required as not all conditions can be considered by pre-trained policies, such as drastic changes of environments or robots by amputation. Hence, meta-learning, or learning to learn is a novel and more promising approach for solving such generic adaptations. Model-based meta-RL has been used in the real-world to adapt the control of a six-leg millirobot to different floor conditions~\cite{GrBAL}. A model-based meta-RL algorithm, FAMLE, was used in the real-world on a Minitaur quadruped, where a latent black-box context vector encoded different environment conditions \cite{FAMLE}.

\begin{figure}[t]
    \centering
    \begin{subfigure}{.2\textwidth}
      \centering
      \includegraphics[width=\textwidth]{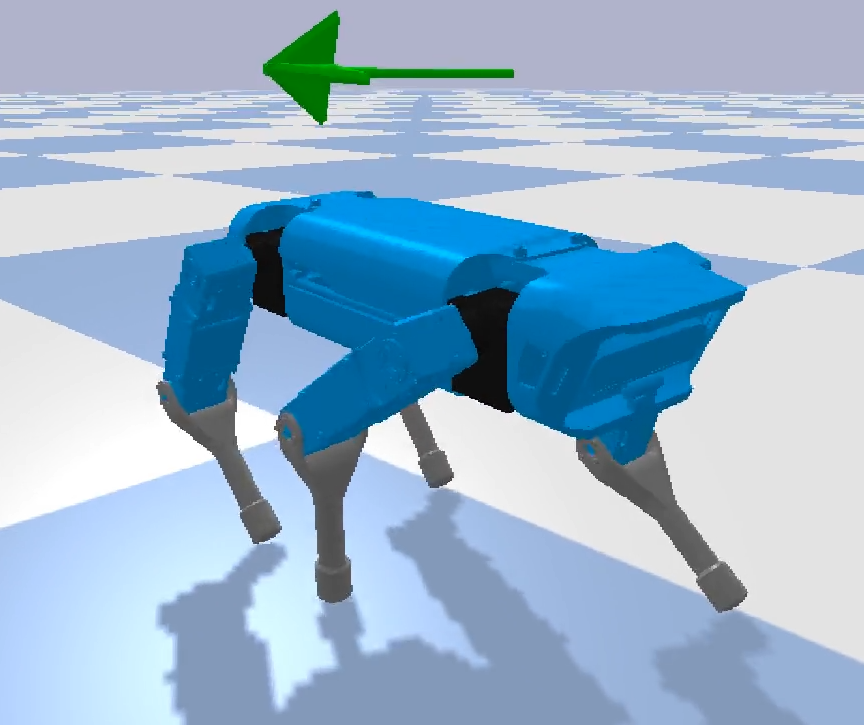}\hfill
      \caption{}
    \end{subfigure}
    \begin{subfigure}{.2\textwidth}
      \centering
      \includegraphics[width=\textwidth]{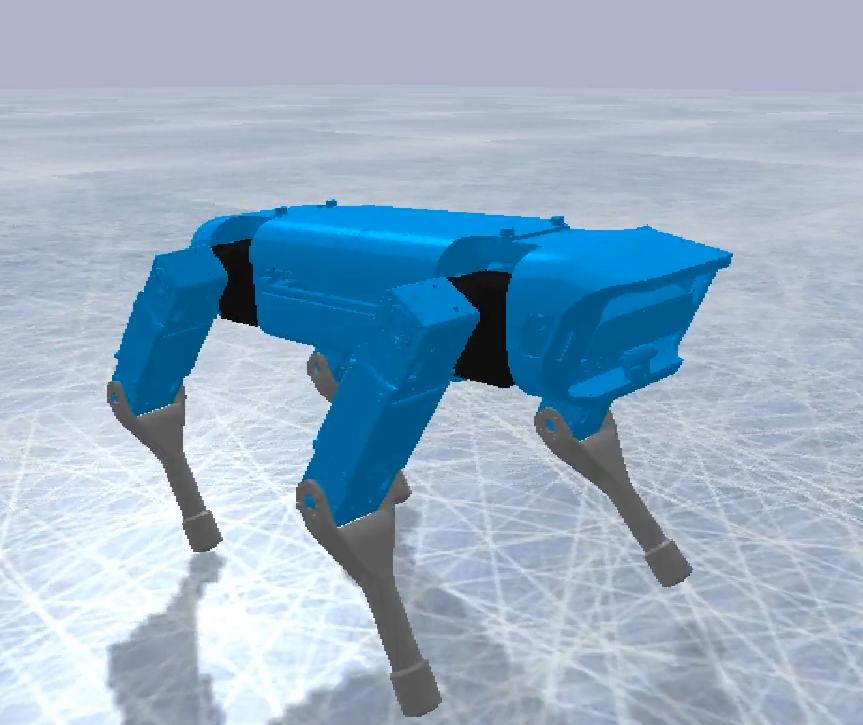}\hfill
      \caption{}
    \end{subfigure}
     \begin{subfigure}{.2\textwidth}
      \centering
      \includegraphics[width=\textwidth]{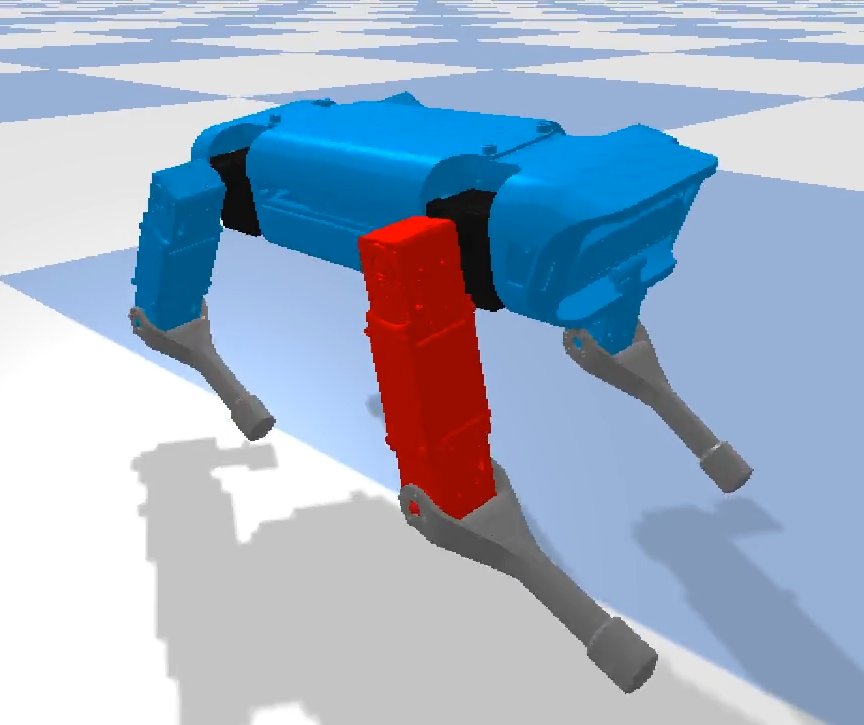}\hfill
      \caption{}
    \end{subfigure}
    \begin{subfigure}{.2\textwidth}
      \centering
      \includegraphics[width=\textwidth]{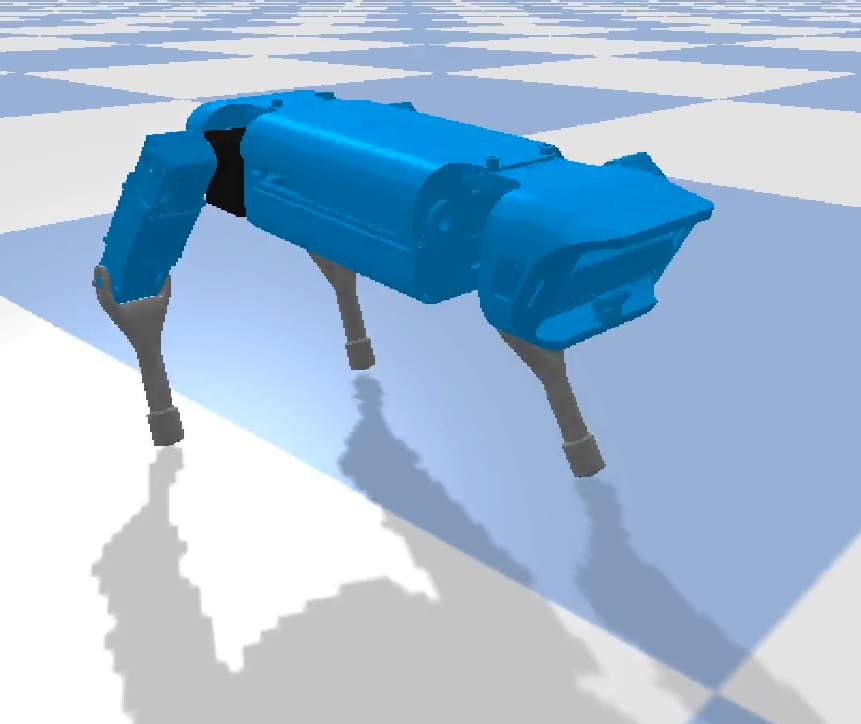}\hfill
      \caption{}
    \end{subfigure}
    \vspace{-2mm}
    \caption{Adaptive and robust locomotion against uncertainties: (a) external force, (b) slippery ground, (c) faulty motors, and (d) leg amputation.}
    \label{fig:quadruped}
    \vspace{-5mm}
\end{figure}

\begin{figure*}[t]
    \centering
    \includegraphics[width=1\linewidth]{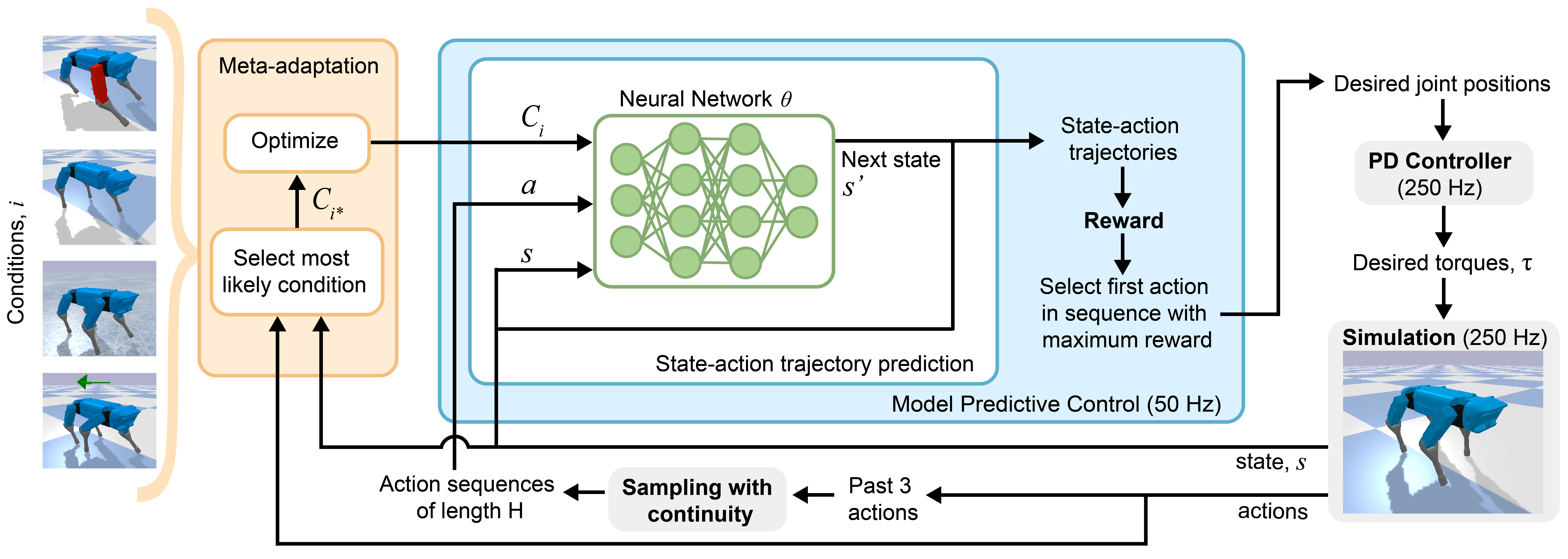}
    \caption{Schematic view of the model-based meta-RL control framework for legged locomotion.}
    \label{fig:schematic}
\end{figure*}


Our proposed method has made new improvements that require no prior knowledge of specific gaits. For example, FAMLE relies on sinusoidal gaits and therefore needs to optimize the amplitudes and phases of sinusoidal patterns by the model predictive control at a low frequency of 0.5Hz. In our work, we directly sample in the joint space at a much higher frequency at 50Hz, and we further improve the sampling process by specifying constraints on velocity, acceleration and jerk of the desired joint trajectories. Our study extensively validated the capability of adaptation in simulated test scenarios with large variations in floor friction, external forces or unexpected damage to joints.

Based on the interaction model, our method allows changing the reward function online and therefore is able to modify the behavior of the robot. For example, the learned controller can track a variable forward velocity, even it has been trained on a fixed desired velocity. Likelihood estimation with condition latent vectors allows the meta-model to adapt to already seen conditions. Meta-training should allow "on the fly" optimization to better adapt to the current unknown condition.

In this paper, we present an improved model-based meta-RL approach to quadruped locomotion that is capable of online adaption to changing environments, as shown in Fig.~\ref{fig:quadruped}. The main contributions of this work are:
\begin{enumerate}
    \item The proposed algorithm is capable of learning from scratch and requires no prior knowledge of the type of gaits, such as periodic phases of leg movements. 
    \item Our methods introduces and applies hard constraints of velocity, acceleration and jerk on the sampled actions during the search process. 
    \item The capability and robustness of online adaption to changes in both the robot and the environment, such as external force disturbances, varying frictions, faulty motors and leg amputation. 
\end{enumerate}


The remainder of the paper is organized as follows. We outline the background in Section~\ref{sec:Background} and related work in Section~\ref{sec:RW}. In Section~\ref{sec:method}, we elaborate the methodology and technical details on the model-based RL algorithm and the improvements by meta-learning. Section~\ref{sec:exp} presents extensive simulation validations, results and analysis. Finally, we conclude and suggest future work in Section~\ref{sec:Conclusion}.

\section{Background}
\label{sec:Background}
This section presents the preliminaries of RL, Model Predictive Control (MPC) and meta-learning. 
\subsection{Reinforcement Learning}
\label{sec:RL}

In reinforcement learning, the agent learns to solve a task in an unknown environment $\mathrm{E}$, defined by a Markov decision process $(\mathrm{S},\mathrm{s}_0, \mathrm{A},\mathrm{f}_\mathrm{E},\mathrm{r})$ where $\mathrm{S}$ is the set of continuous states of the environment $\mathrm{E}$, $\mathrm{s}_0$ the initial state, $\mathrm{A}$ the set of continuous actions the agent can perform in the environment, $\mathrm{f}_\mathrm{E}: \mathrm{S}\times \mathrm{A}\times\mathrm{S} \rightarrow \mathbb{R}$ the probabilistic transition function and $\mathrm{r}: \mathrm{S}\times \mathrm{A} \rightarrow \mathbb{R}$ the reward function. 

The goal of the agent is to learn a policy $\Pi_{\theta}: \mathrm{S} \rightarrow \mathrm{A}$, parameterized by $\theta$, which decides which action to perform given the current state $\mathrm{s}_\mathrm{t}$ to \textit{maximize the long-term reward} $\mathrm{R}(\mathrm{t})=\underset{\mathrm{i}=\mathrm{t}}{\overset{\mathrm{t}+\mathrm{H}}{\sum}}\gamma^{\mathrm{i}-\mathrm{t}}\mathrm{r}(\mathrm{s}_\mathrm{i},\Pi(\mathrm{s}_\mathrm{i}))$, where $\mathrm{H}$ is the horizon and $\gamma$ is the discount factor. 

Model-free RL focuses in directly learning such a policy, whereas model-based RL focuses on learning a model of the transition function -- the transition of states given the current state and actions  -- which can be used to train the policy with fictive transitions or with model predictive control.

\subsection{Model Predictive Control}
\label{sec:MPC}
Given the current state $\mathrm{s}_\mathrm{t}$ and a horizon $\mathrm{H}$, Model Predictive Control (MPC) uses a forward model of dynamics to select an action sequence $\mathrm{a}_{\mathrm{t}:\mathrm{t}+\mathrm{H}}$ which maximizes the predicted cumulative reward $\mathrm{R}(\mathrm{t})$. The agent performs the first action $\mathrm{a}_\mathrm{t}$ from the action sequence and collects the resulting state $\mathrm{s}_{\mathrm{t}+1}$. The MPC then repeats such optimization and allows the agent to alleviate the possible error in the model prediction. Compared to model-free RL, we can change the reward \textit{online} to control the agent's behavior using model-based RL in an MPC fashion. 

\subsection{Meta-Learning}
\label{sec:ML}
We use meta-learning to train an agent to solve several tasks, where the neural network learns to adapt to several varying conditions, such as different floor frictions, the presence of external disturbances or having a damaged motor. For the neural network model, the initial set of weights $\theta^*$ must be found, such that only a small number of gradient descent steps with little collected data in a unknown environment can produce effective adaptations. 

\section{Related Work}
\label{sec:RW}
\subsection{Model-Free Deep Reinforcement Learning}
Proximal Policy Optimization~\cite{PPO} has been used to train a model-free controller to control Minitaur in simulation, producing trotting and gallop gaits \cite{SimToReal} on the real robot. Soft Actor-Critic (SAC)~\cite{SAC} has also been used to train on the real Minitaur robot within two hours~\cite{LearningToWalk, ha_learning_2020}. This limitation of sampling-efficiency motivates us to focus on model-based RL. 

\subsection{Model-Based Deep Reinforcement Learning}
There are three types of model-based RL: learning to predict the expected return from a starting state distribution, for example using Bayesian Optimization~\cite{BayesianOptimisation}; learning to predict the outcome from a given starting state and given policy~\cite{noveltySearch, MapElite, IMGEP}; and learning to model the transition function using a forward dynamical model. Here, we use the third type of model.

Forward models of dynamics are either deterministic or probabilistic, where deterministic models can be linear models~\cite{levineLinearModel} or neural networks~\cite{WorldModel}, and probabilistic models estimate uncertainty for modeling stochastic environments or estimating the long-term prediction uncertainty. Gaussian Processes~\cite{Pilco, Multi-dex} or Bayesian neural networks~\cite{Gal2015DropoutB} can be used to scale the abilities of Gaussian Processes models to higher dimensional environments. 

For locomotion tasks, model-based RL with a forward model can have the same performance as model-free methods, while requiring at least an order of magnitude less samples~\cite{Chua}. An ensemble of feed-forward neural networks is used to model the forward dynamics of the environment with uncertainty estimation. MPC uses this uncertainty estimation to formulate a more robust control which alleviates early overfitting model-based RL. The same method has been used with meta-learning to adapt the control of a 6-leg real millirobot to different floors \cite{GrBAL}. 

\subsection{MPC and Meta-Learning}
Several optimization methods have been used for model-based RL, for example, Model Predictive Path Integral~\cite{MPPI}, random shooting \cite{LearningToWalk} or Cross-entropy method~\cite{CEM, Chua}. We use random shooting for the simplicity, easy parallelism and proven performances on real robots~\cite{GrBAL}.  

There are two main methods: a meta-learner model outputs the set of initial weights  $\theta^*$ of the learner \cite{learningToLearn}, or $\theta^*$ is optimized using a meta-loss, it can be gradient-descent~\cite{MAML} or evolutionary strategies~\cite{ES-ML}. Gaussian processes have been used~\cite{ml-gp} but only for low dimension environments. Meta-RL has been used with model-free RL~\cite{noRML}, model-based RL~\cite{GrBAL} or a mix of both~\cite{M3PO}. For model-based RL, gradient based meta-learning was shown to be more data-efficient, resulting in a better and faster adaptation~\cite{GrBAL}. Hence, we use gradient-based meta-learning. 

For increasing generalization and adaptation to unseen condition of the environment, an adversarial loss has been used \cite{adMRL}. Other methods employ context variables \cite{context-variable}, bias transformation \cite{bias-transformation}, or condition latent vector~\cite{FAMLE}, to learn different input of sub-parts of the model for different condition, and then adapt this sub-parts to the current condition. 


\section{Methodology}
\label{sec:method}
This section presents details of the model-based RL algorithm as discussed in Section~\ref{sec:MBRL} and meta-learning algorithm Section~\ref{sec:meta-learning}. We highlight our improvements which results in new robot capability of robust and versatile walking without a predefined, parameterized gait.

\subsection{Model-Based Reinforcement-Learning algorithm}
\label{sec:MBRL}
The model-based RL algorithm runs at 50Hz, sending desired actions to PD controllers running at 250Hz to generate torques for physics simulation. The algorithm is composed of two main parts: the forward dynamics model, and MPC. Fig.~\ref{fig:schematic} illustrates the schematics of the control framework. 

\subsubsection{The Forward Model of Dynamics}
We use a fully-connected feed-forward neural network, with two hidden layers of 256 units using a ReLU (Rectified Linear Unit) activation function. It takes the concatenation of the current state and action ($\mathrm{s}_{\mathrm{t}},\mathrm{a}_\mathrm{t}$) as input, and learns to predict the difference in the resulting state: $\Delta \mathrm{s}_\mathrm{t}= \mathrm{s}_\mathrm{t+1}-\mathrm{s}_{\mathrm{t}}$, which is a standard means to get the prediction $\mathrm{s}_\mathrm{t+1} = \mathrm{s}_{\mathrm{t}} + \Delta \mathrm{s}_\mathrm{t}$. 

The model parameter $\theta$ is the set of weights of the connections between the units. It is optimized using the gradient-based optimizer Adam~\cite{adam} on a dataset of triplet $\mathrm{D}\in S\times A \times S$ using mean squared error as loss function. We depict details of the model-based RL algorithm in Algorithm~\ref{algo:MBRL}.

Compared to the work in~\cite{FAMLE} for the Minitaur, our study formulate the state space as: the angular joints positions and velocities, the base orientation angles and angular rates, and the linear base velocities. The addition of the angular velocity of the base is the key of our success for controlling the robot at 50Hz. In contrast, only Euler angles and angular rates of the base in the horizontal plane were used in~\cite{FAMLE} to control the Minitaur gait parameters at a much lower frequency of 0.5Hz. 

\begin{algorithm}[t]
\caption{Model-based reinforcement learning algorithm \newline Input: Environment $\mathrm{E}$.}
\label{algo:MBRL}
\begin{algorithmic}
\STATE $\mathrm{D}$ = $\mathrm{N}_{init}$ episodes in $\mathrm{E}$ with a random controller
\STATE $\theta=$ random weights
\FOR{$\mathrm{N}_\text{training}$ episodes}
    \STATE Train $\mathrm{M}_\theta$ on $\mathrm{D}$ using Adam
    \STATE $\mathrm{s}_0=$ $\mathrm{E}$.reset()
    \FOR{$\mathrm{t}=0$ to $\mathrm{T}-1$ steps}
        \STATE $\mathrm{a}_\mathrm{t}=$ MPC($\mathrm{M}_\theta$, $\mathrm{s}_{t}$, $\mathrm{a}_{\mathrm{t}-3:\mathrm{t}-1}$)
        \STATE $\mathrm{s}_\mathrm{t+1}=$ $\mathrm{E}$.step($\mathrm{a}_\mathrm{t}$)
        \STATE $\mathrm{D}=\mathrm{D}\cup \{( \mathrm{s}_{\mathrm{t}}, \mathrm{a}_\mathrm{t}, \mathrm{s}_\mathrm{t+1})\}$
    \ENDFOR
\ENDFOR
\end{algorithmic}
\end{algorithm}

\subsubsection{Model Predictive Control}
The method of random shooting is implemented which is suitable for parallel computing, and the algorithm is detailed in Algorithm~\ref{algo:MPC}. At each time step, $\mathrm{N}_\text{pop}$ action sequences of length $\mathrm{H}$ are sampled. Each sequence is evaluated starting with the current state, using the model to estimate the corresponding state trajectory. From these trajectories, long term reward is computed and the action with the highest estimated reward is selected.

Real actuators have inherent limitations in velocity, acceleration and jerk. Instead of uniformly sampling desired joint angles within the limits, continuity constraints are used, where each desired joint state of the sequence is sampled using previous joint positions to ensure velocities, accelerations and jerks are smooth and bellow their respective limits. 

As the improvement to the previous work~\cite{FAMLE}, we enforced physical constraints during sampling of actions: $\mathrm{q}_{min}\le\mathrm{q}\le\mathrm{q}_{max}$, $|\mathrm{A}|\le \mathrm{A}_{max}$, $|\mathrm{V}|\le \mathrm{V}_{max}$ and $|\mathrm{J}|\le \mathrm{J}_{max}$, where $\mathrm{q}$, $\mathrm{V}$, $\mathrm{A}$ and $\mathrm{J}$ are the desired joint angle, velocity, acceleration and jerk, respectively. The limits of velocity, acceleration and jerk are the soft constraints for the smoothness and continuity of actions. For safety reasons, regarding the joint position limits, we further imposed hard constraints of sampled actions on $\mathrm{q}$ to avoid hitting the physical limit of joint movements. 

This improvements on sampling enforces the MPC in a more suitable subspace. During training, it increased the distance traveled compared to the default condition during a 10s episode by an order of 2: from $1.64\pm0.12$m (without), to $3.16\pm0.12$m (with), where p-value $\le$ 0.001 on 20 episodes. It also reduces the observed jerk by an order of 5: from $1.3\times {10}^{4} \pm 260$ rad/$\mathrm{s}^3$ (without) to $2.4 \times 10^3 \pm 60$ rad/$\mathrm{s}^3$ (with), where p-value $\le$ 0.001 on 20 episodes.

\begin{algorithm}[t] 
\caption{MPC algorithm \newline Inputs: A model $\mathrm{M}_\theta$, initial state $\mathrm{s}_0$, past actions $\mathrm{a}_{-3:-1}$, horizon $\mathrm{H}$ and discount factor $\gamma$.}
\label{algo:MPC}
\begin{algorithmic}
\STATE Sample $\mathrm{a}_{1:\mathrm{H}}^{1:\mathrm{N}_\text{pop}}$ using $\mathrm{a}_{-3:-1}$ for continuity
\FOR{$\mathrm{i}=1$ to $\mathrm{N}$}
    \STATE $\mathrm{R}^{\mathrm{i}} = 0$
\ENDFOR
\FOR{$\mathrm{t}=1$ to $\mathrm{H}$ steps}
    \FORALL{$\mathrm{i}=1$ to $\mathrm{N}_\text{pop}$ samples}
        \STATE $\mathrm{s}^{\mathrm{i}}_\mathrm{t+1}=\mathrm{M}_\theta(a_{\mathrm{t}}^{\mathrm{i}}, \mathrm{s}_{\mathrm{t}})$
        \STATE $\mathrm{R}^\mathrm{i} = \mathrm{R}^\mathrm{i} + \gamma ^{\mathrm{t}-1}\mathrm{r}(\mathrm{s}^{\mathrm{i}}_{\mathrm{t}}, \mathrm{a}^\mathrm{i}_{\mathrm{t}})$
    \ENDFOR
\ENDFOR
\STATE $\mathrm{i}^*=\underset{\mathrm{i}}{\mathrm{argmax}}$ $\mathrm{R}^\mathrm{i}$
\RETURN $\mathrm{a}_1^{\mathrm{i}^*}$
\end{algorithmic}
\end{algorithm}

\subsection{Meta-learning algorithm}
\label{sec:meta-learning}
Before meta-training, an expert is trained for each training condition using the proposed model-based RL algorithm to collect its training data. To adapt the model to each condition $\mathrm{i}$, a specific latent vector $\mathcal{C}_\mathrm{i}$ is optimized during meta-learning using the regression loss on the data of the corresponding condition. This vector of fixed dimensions is then given to the input layer, alongside the current state and action when the condition $\mathrm{i}$ is selected. We use a first order meta-learning called Reptile~\cite{reptile}, which is composed of two phases: meta-training (Algorithm~\ref{algo:meta_training}) and meta-adaptation (Algorithm~\ref{algo:meta_adaptation}). 

\subsubsection{Meta-training}

\begin{algorithm}[t] 
\caption{Meta-learning training, called once before adaptation \newline Inputs: $\mathrm{N}$ datasets $\mathrm{D}_{0:\mathrm{N}-1}$ from $\mathrm{N}$ different conditions.}
\label{algo:meta_training}
\begin{algorithmic}
\STATE $\theta^*$ = random weights
\FOR{$\mathrm{i}=0$ to $\mathrm{N}-1$}
    \STATE $\mathcal{C}_{\mathrm{i}}$ = random vector
\ENDFOR
\FOR{$\mathrm{n}_{outer}=0$ to $\mathrm{N}_{outer}-1$}
    \STATE $\mathrm{i} = \mathrm{n}_{outer} \mod \mathrm{N}$
    \STATE $\theta = \theta^*$
    \FOR{$\mathrm{n}_{inner}=1$ to $\mathrm{N}_{inner}$}
        \STATE $\mathcal{L} = \tfrac{1}{|\mathrm{D}_{\mathrm{i}}|}\underset{(\mathrm{s},\mathrm{a},\mathrm{s}')\in \mathrm{D}_\mathrm{i}}{\sum} (\mathrm{M}_\theta(\mathrm{s},\mathrm{a}, \mathcal{C}_{\mathrm{i}})-\mathrm{s}')^2$ 
        \STATE $\theta= \theta - \alpha \nabla_\theta \mathcal{L}$ 
        \STATE $\mathcal{C}_{\mathrm{i}}= \mathcal{C}_{\mathrm{i}} - \alpha \nabla_{\mathcal{C}_{\mathrm{i}}} \mathcal{L}$ 
    \ENDFOR
    \STATE $\theta^* = \theta^* + (1-\tfrac{\mathrm{n}_{outer}}{\mathrm{N}_{outer}})\beta (\theta-\theta^*)$ 
\ENDFOR
\RETURN $\theta^*$
\end{algorithmic}
\end{algorithm}

\begin{algorithm}[t] 
\caption{Meta-learning adaptation, called at each step \newline Inputs: a meta-trained set of weights $\theta^*$, a list of learned $\mathrm{N}$ condition latent vectors $\mathcal{C}_{1:\mathrm{N}}$, a dataset $\mathrm{D}$ of the past $\mathrm{K}$ steps.}
\label{algo:meta_adaptation}
\begin{algorithmic}

\FOR{$\mathrm{i}=1$ to $\mathrm{N}$}
    \STATE $\mathcal{L}_\mathrm{i} = \tfrac{1}{|\mathrm{D}|}\underset{(\mathrm{s},\mathrm{a},\mathrm{s}')\in \mathrm{D}}{\sum} (\mathrm{M}_\theta^*(\mathrm{s},\mathrm{a}, \mathcal{C}_\mathrm{i})-\mathrm{s}')^2$
\ENDFOR
\STATE $i^*= \underset{\mathrm{i}}{\operatorname{argmin}} \mathcal{L}_i$
\STATE $\theta = \theta^*$
\FOR{$n_{inner}=1$ to $\mathrm{N}_{inner}$}
    \STATE $\mathcal{L} = \tfrac{1}{|\mathrm{D}|} \underset{(\mathrm{s},\mathrm{a},\mathrm{s}')\in D}{\sum} (M_\theta(\mathrm{s},\mathrm{a}, \mathcal{C}_{\mathrm{i^*}})-\mathrm{s}')^2$
    \STATE $\theta = \theta - \alpha \nabla_\theta \mathcal{L}$ 
    \STATE $\mathcal{C}_{\mathrm{i^*}} = \mathcal{C}_{\mathrm{i^*}} - \alpha \nabla_{\mathcal{C}_{\mathrm{i^*}}} \mathcal{L}$ 
\ENDFOR
\RETURN $\theta$, $\mathcal{C}_{\mathrm{i^*}}$
\end{algorithmic}
\end{algorithm}

The initial set of weights $\theta^*$ and each condition latent vector $\mathcal{C}_\mathrm{i}$ are optimized for adaptation. Meta-training is separated into two nested loops. In the inner loop, one training dataset $\mathrm{D}_{\mathrm{i}}$ and its corresponding condition latent vector $\mathcal{C}_\mathrm{i}$ are selected. The model weights $\theta$ are initialized to $\theta^*$ and Adam~\cite{adam} optimize both of them for the regression loss of the current dataset $\mathrm{D}_{\mathrm{i}}$.

In the outer loop, $\theta^*$ is optimized by taking a small step, with a linearly decreasing schedule, towards the optimized weights of the inner-loop. This allows $\theta^*$ to converge to a nearby point (in the euclidean sense) to the optimal set of weights of each training condition. We detail the algorithm in Algorithm~\ref{algo:meta_training}.

\subsubsection{Meta-adaptation}
At each time step, we select the most likely training condition using the previous $\mathrm{K}$ time steps, each condition latent vector and the set of weights $\theta^*$. 
We then optimize the corresponding latent vector and the set of model weights, starting from $\theta^*$, using the same optimization procedure as the inner loop but with the past $\mathrm{K}$ steps. We detail the algorithm in Algorithm~\ref{algo:meta_adaptation}.

After the set of weights and the condition latent vector are optimized for the current condition, we use the MPC to select the optimal action to apply, and then new state information is collected, and the whole meta-adaptation iterates. This procedure allows any changes in the condition to be detected, and therefore the agent can adapt accordingly. 

\subsection{Limitations}
Apart from the standard classical robot control of tuning PD gains and joints limits, the proposed method still requires fine-tuning of reward function, model architecture, hyper-parameters for the meta-learning and the adaptation. The use of MPC instead of a neural network policy has a trade-off between real-time computation and performance, i.e. MPC performs better in terms of adaption but requires more computation needed from the sampling procedure.

\begin{figure}[t]
    \centering
      \begin{subfigure}{\linewidth}
      \centering
      \includegraphics[width=\linewidth]{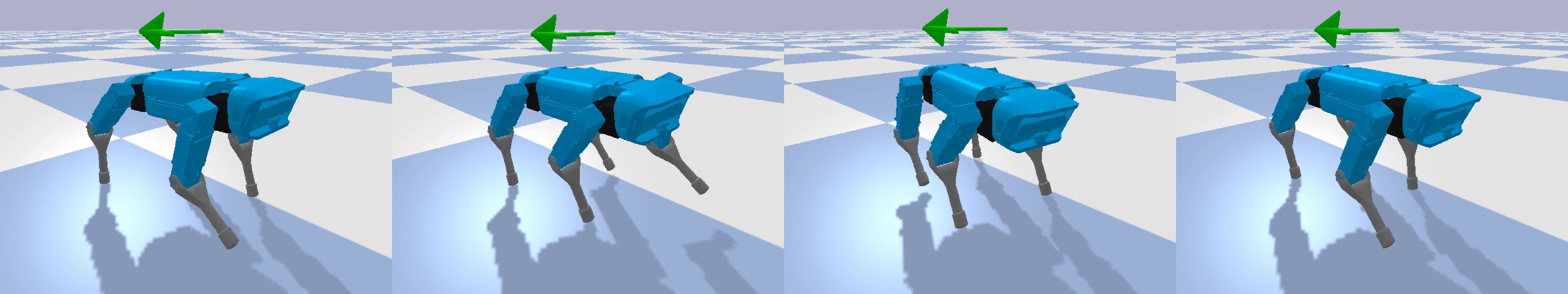}
      \caption{}
    \end{subfigure}
    \begin{subfigure}{\linewidth}
      \centering
      \includegraphics[width=\linewidth]{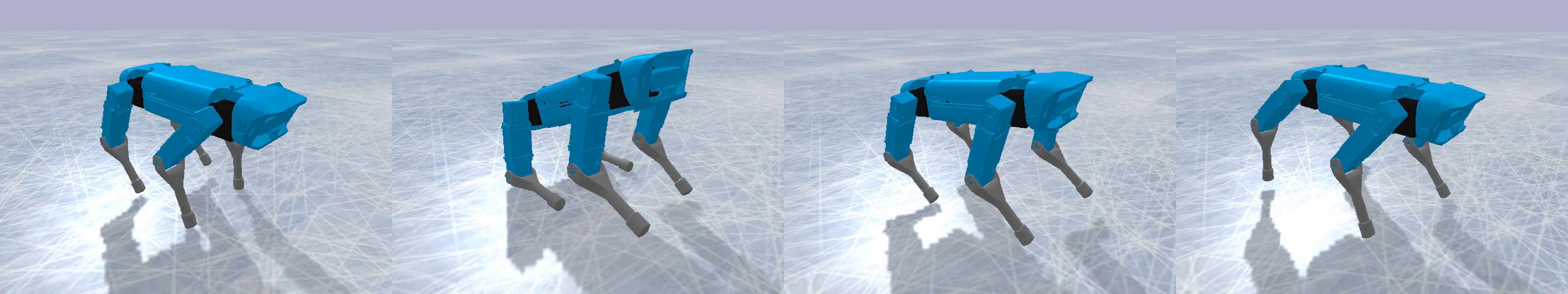} 
      \caption{}
    \end{subfigure}
        \begin{subfigure}{\linewidth}
      \centering
      \includegraphics[width=\linewidth]{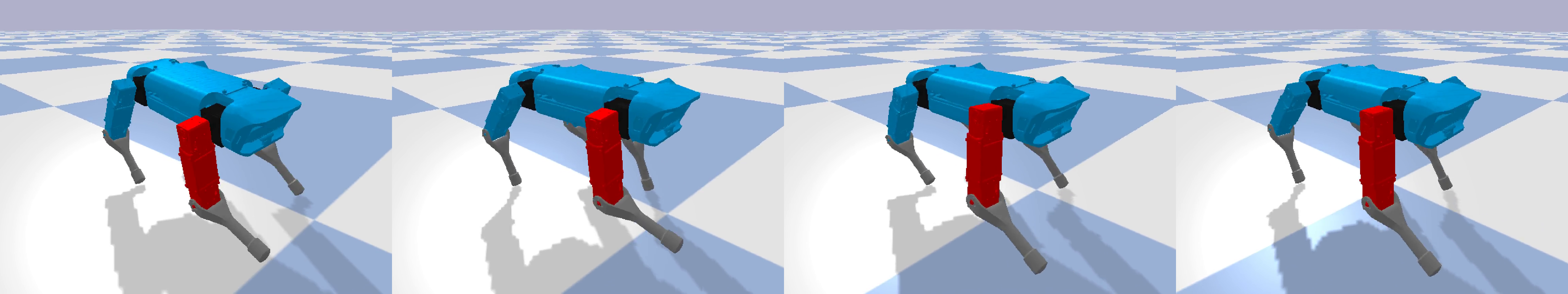}
      \caption{}
    \end{subfigure}
    \begin{subfigure}{\linewidth}
      \centering
      \includegraphics[width=\linewidth]{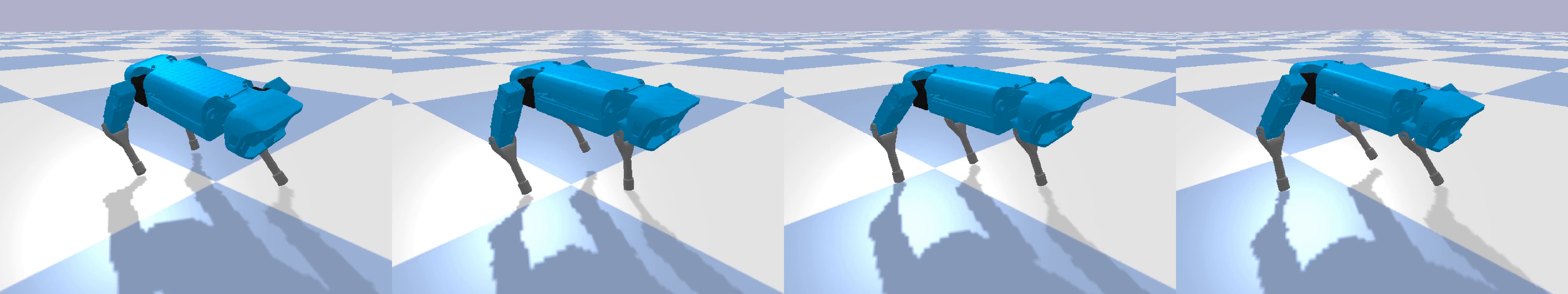}
      \caption{}
    \end{subfigure}
    \caption{Walking in presence of large uncertainties: (a) constant external force disturbance, (b) low-friction slippery ground, (c) faulty motors, and (d) with one missing leg.}
    \vspace{-0mm}
    \label{fig:snapshots}
\end{figure}

\section{Results}
\label{sec:exp}
We used a custom version of the robot model (adapted from the open source SpotMicro robot~\cite{kim_2019}) in PyBullet simulation to validate our method. Here, we first present the learning capability of the model-based RL algorithm on SpotMicro with a first adaption to a sequence of different conditions (Section~\ref{sec:general_results}). We further validate the adaptation capability of the proposed meta-learning under various fixed frictions (Section~\ref{sec:fixed_friction}) and time-varying, decreasing friction (Section~\ref{sec:decreasing_friction}).

\subsection{Overview}
\label{sec:general_results}

\begin{figure}[t]
    \centering
      \begin{subfigure}{\linewidth}
      \centering
      \includegraphics[width=\linewidth]{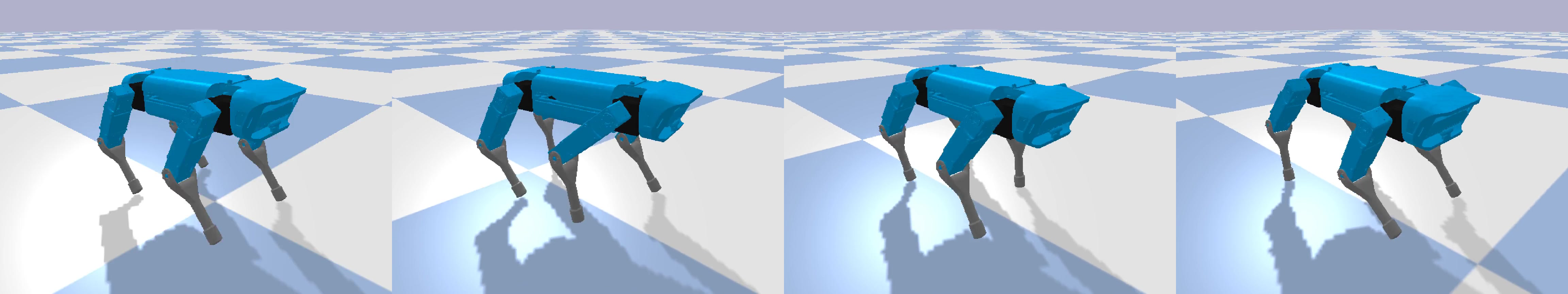}
      \caption{}
    \end{subfigure}
    \begin{subfigure}{\linewidth}
      \centering
      \includegraphics[width=\linewidth]{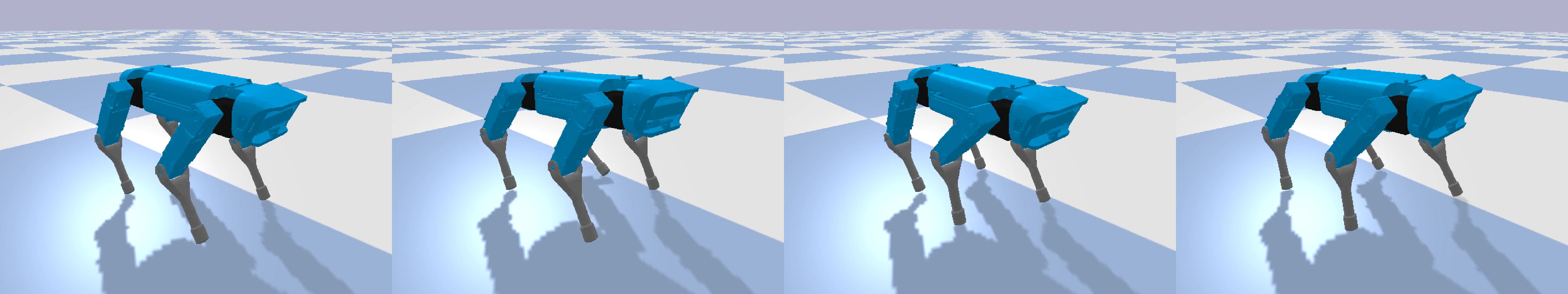}
      \caption{}
    \end{subfigure}
        \begin{subfigure}{\linewidth}
      \centering
      \includegraphics[width=\linewidth]{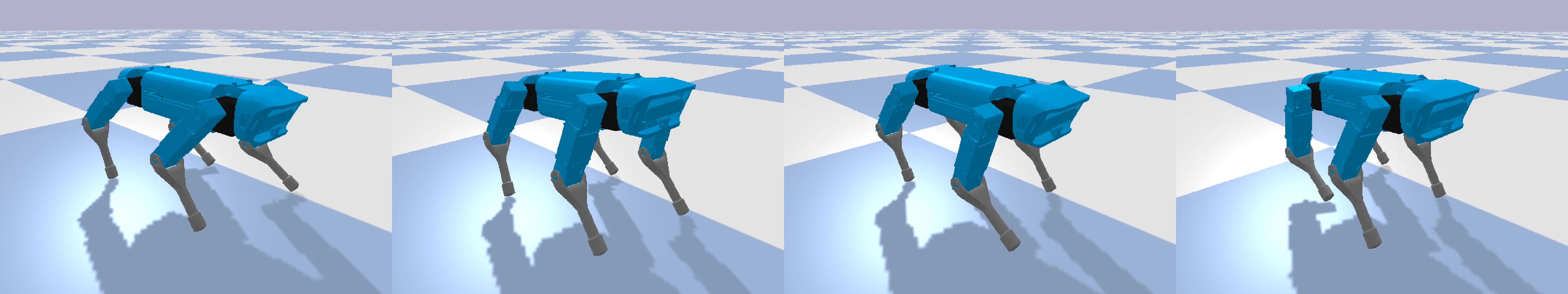}
      \caption{}
    \end{subfigure}
    \caption{Walking with a continuously changing velocity from 0.5m/s to -0.2m/s: (a) forward, (b) static, and (c) backward.}
    \vspace{-0mm}
    \label{fig:varying_velocity_snapshots}
\end{figure}

\begin{figure*}[t]
    \centering
    \includegraphics[width=\linewidth]{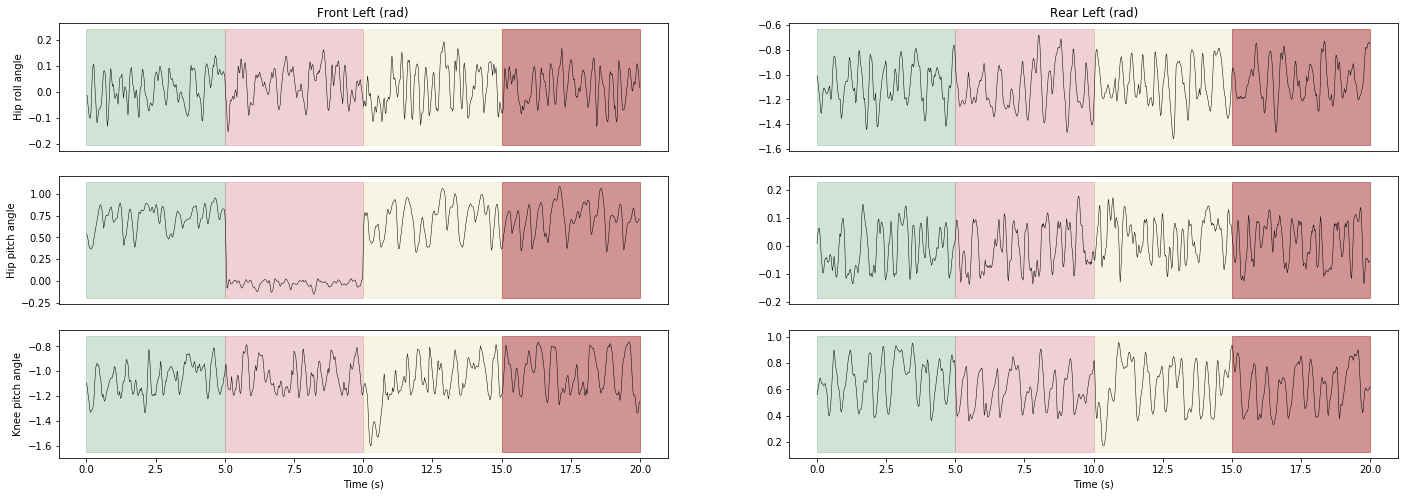} 
    \caption{Measured joints trajectories of SpotMicro from a 20s test scenario where the robot or environment changed every 5s: default friction $\mu=0.8$ (\textcolor{YellowGreen}{\textbf{green}}), blocked right hip pitch joint (\textcolor{Lavender}{\textbf{red}}), slippery ground (\textcolor{Dandelion}{\textbf{yellow}}) and constant external push (\textcolor{Bittersweet}{\textbf{brown}}).}
    \vspace{-2mm}
    \label{fig:joints_trajectories}
\end{figure*}

\begin{table*}[t]
\centering
\begin{tabular}{|l|c|c|c|c|}
\hline
Expert\textbackslash{}Condition & Default           & Slippery         & Lateral Force     & Damaged Motor      \\ \hline
Default                         & 100\%, 3.2$\pm$ 0.2 & 20\%, 0.7$\pm$ 0.7 & 40\%, 2.0$\pm$ 1.0  & 100\%, 0.7$\pm$ 0.2  \\ \hline
Slippery                        & 30\%, 1.3$\pm$ 0.5  & 90\%, 2.4$\pm$ 0.4 & 10\%, 0.7$\pm$ 0.4  & 100\%, 0.4$\pm$  0.1 \\ \hline
Lateral Force                   & 0\%                & 0\%                & 70\%, 2.3 $\pm$ 0.7 & 90\%, 0.4$\pm$ 0.1   \\ \hline
Damage Motor                    & 90\%, 0.5$\pm$ 0.2  & 70\%, 0.3$\pm$ 0.2 & 30\%, 0.6$\pm$ 0.3  & 100\%, 2.5$\pm$ 0.1  \\ \hline
Meta-Trained                    & 70\%, 3.0$\pm$ 0.6  & 80\%, 2.4$\pm$ 0.7 & 40\%, 2.0$\pm$ 1.3  & 100\%, 2.7$\pm$ 0.1  \\ \hline
\end{tabular}
\caption{Success rate and average distance of travel for 10 episodes from different expert and the meta-trained model.}
\label{tab:success_rates}
\end{table*}

We trained the expert model with a default condition of friction $\mu=0.8$, and this resulted default controller for walking is robust to perturbations, withstanding several pushes of 10N for 0.2s. After 300 of 10s-episodes which produced training data in the given condition, the quadruped was able to walk on slippery ground (friction coefficient $\mu = 0.2$), against external forces or with a blocked motor or a missing/amputated leg. Tab.~\ref{tab:success_rates} shows the comparison between experts and meta-trained models under different conditions.

Using the proposed meta-learning method, the agent was able to adapt to four different conditions: default, fixed front-right hip motor, slippery ground and external forces. It traveled an average distance of $5.1\pm1.5$m compared to a default expert which traveled $2.6\pm0.8$m (averaged over 20 episodes). The joint trajectories from these test scenarios are shown in Fig.~\ref{fig:joints_trajectories}.

\begin{figure}[t]
    \centering
    \includegraphics[width=\linewidth]{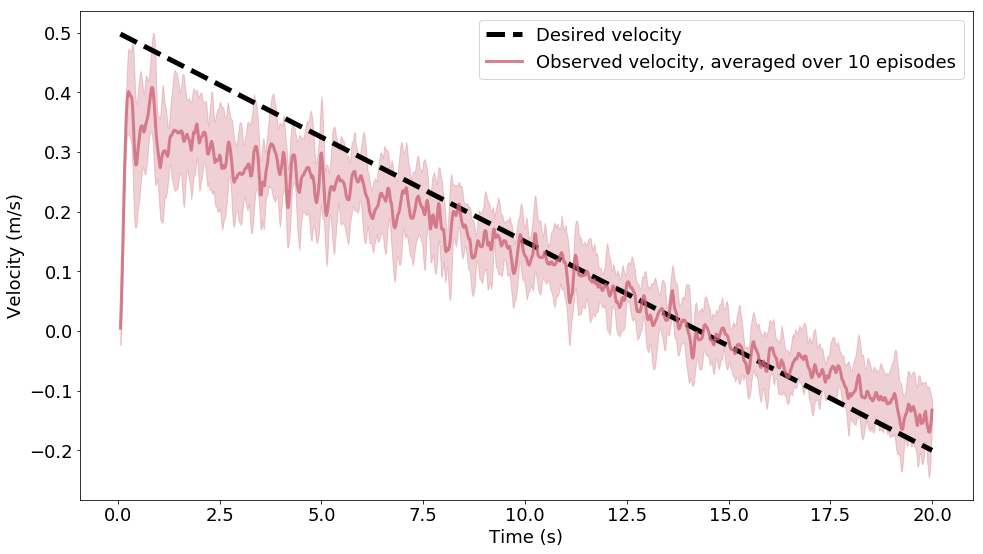}
    \caption{Online tracking of continuous and variable walking velocity using the model trained only at the fixed velocity of 0.5m/s, while the desired velocity in the reward function changes from from 0.5m/s to -0.2m/s during the test scenario.}
    \vspace{-2mm}
    \label{fig:varying_velocity}
\end{figure}

\begin{figure}[t]
    \centering
    \includegraphics[width=\linewidth]{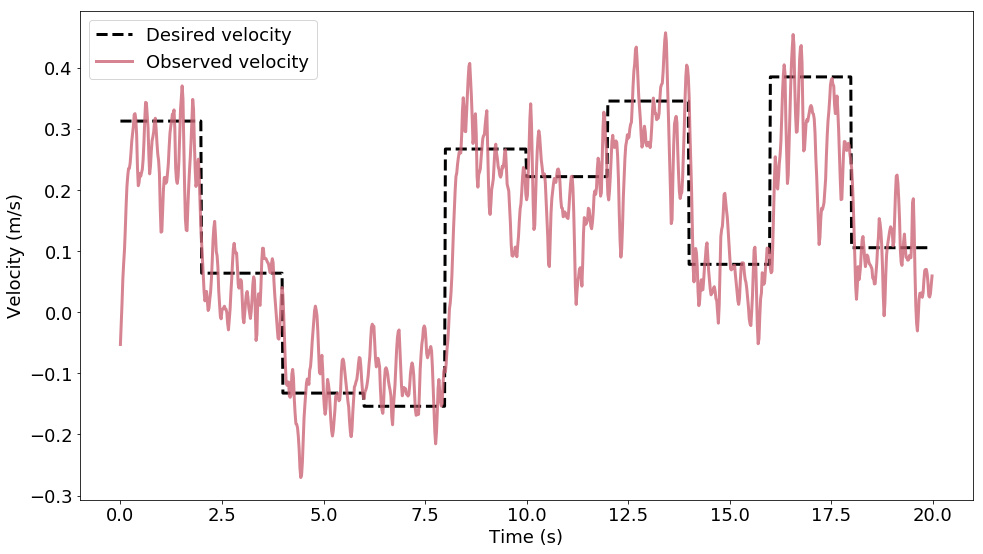}
    \vspace{-4mm}
    \caption{Online tracking of discrete and variable walking velocity using the model trained only at the fixed velocity of 0.5m/s, while every 2 s the desired velocity is randomly sampled within 0.5m/s and -0.2m/s during the test scenario.}
    \vspace{-0mm}
    \label{fig:random_velocity}
\end{figure}

The controller can achieve variable walking speed, despite being trained with only at a constant desired forward velocity, we can command different desired velocities online continuously, as shown in Fig.~\ref{fig:varying_velocity_snapshots} and Fig.~\ref{fig:varying_velocity}. Moreover, the trained expert controller can also generate continuous control actions to track discrete, discontinuous commanded velocities (see Fig. \ref{fig:random_velocity}).

\begin{figure}[t]
    \centering
    \includegraphics[width=\linewidth]{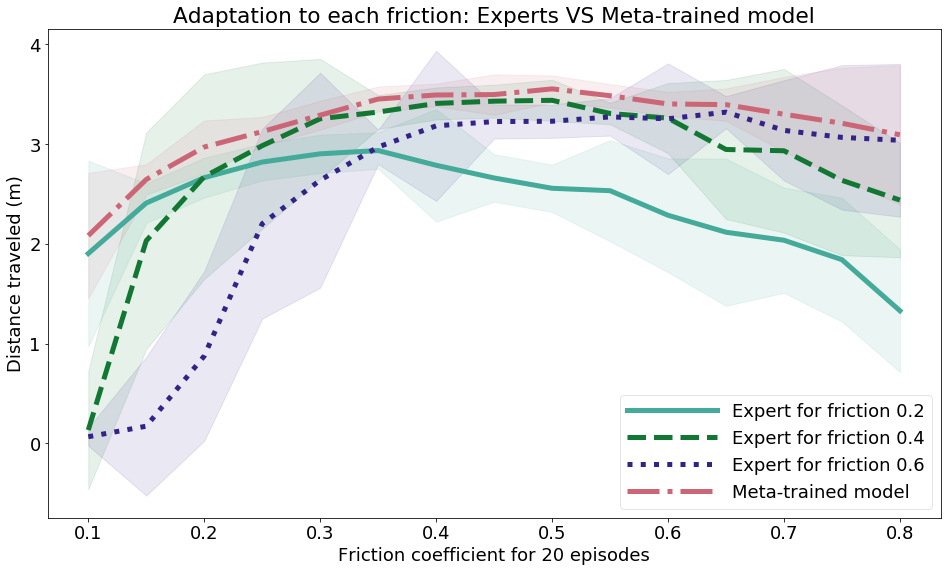}
    \vspace{-0mm}
    \caption{Distance traveled for the full range of frictions from 3 expert models and the meta-trained model.}
    \vspace{-2mm}
    \label{fig:fixed_friction_adaptation}
\end{figure}

\begin{figure}[t]
    \centering
    \includegraphics[width=\linewidth]{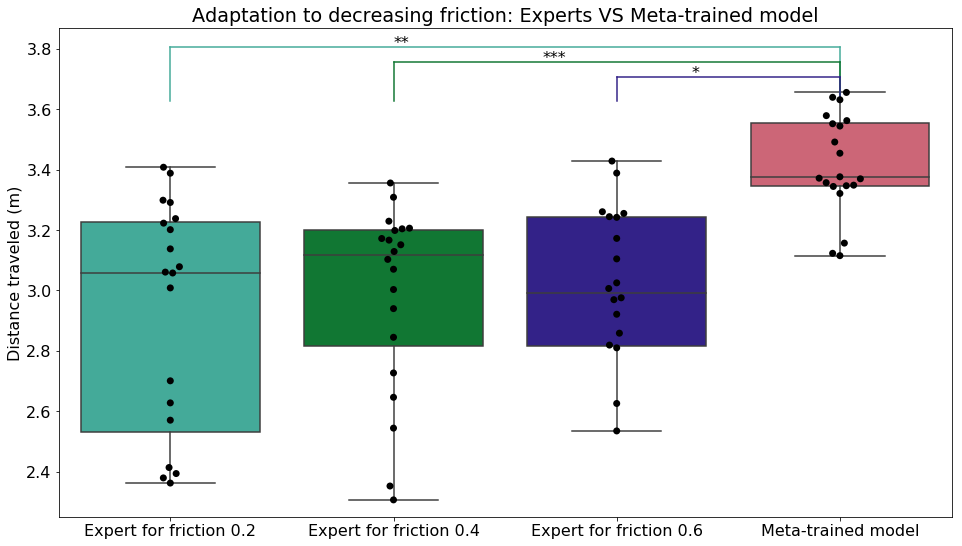}
    \vspace{-0mm}
    \caption{Distance traveled with decreasing friction from the 3 expert models and the meta-trained model (p-value: $*<0.05$, $**<0.01$ and $***<0.001$). }
    \vspace{-2mm}
    \label{fig:changing_friction_adaptation}
\end{figure}

\subsection{Ground with Constant Friction}
\label{sec:fixed_friction}
We evaluated the adaptation capability using the meta-trained model and compared it to experts over the full range of different frictions (0.1 to 0.8 with 0.05 increments). We first trained 5 sets of experts for frictions 0.2, 0.4 and 0.6, using 300 10s-episodes, i.e., 50 minutes of data. Then we meta-trained 5 meta-models to adapt to these 3 frictions, each using one set of experts data, with the purpose that they could adapt to the full range afterwards.

Each set of these 5 models was evaluated for each friction with 4 10s-episodes, so this gives 20 evaluations per expert and 20 evaluations for the meta-learning. The meta-trained models outperformed the experts on the full range of frictions, see Fig.~\ref{fig:fixed_friction_adaptation}. As expected, each expert had its best performance when the friction constant is around its trained value.

\subsection{Ground with Decreasing Friction}
\label{sec:decreasing_friction}j

As a comparison, we used the same experts and meta-model and benchmarked thei adaptation capability on a ground with continuously decreasing friction. We evaluated each set of models with 4 10s-episodes where friction coefficient $\mu$ started at 0.8 and linearly decreased to 0.1. This gave 20 evaluations per expert and 20 evaluations for meta-learning.

\begin{figure}[t]
    \centering
    \includegraphics[width=\linewidth]{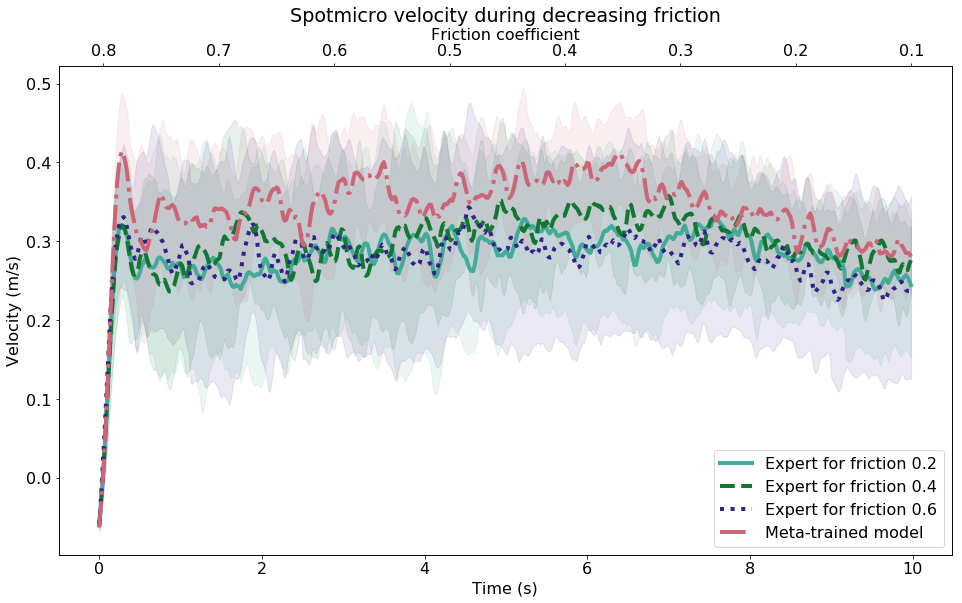}
    \vspace{-4mm}
    \caption{Forward walking velocity with decreasing friction from 3 expert models and the meta-trained model.}
    \vspace{-0mm}
    \label{fig:velocity}
\end{figure}

The meta-trained models demonstrated better walking performance and traversed farther (3.38m) than the experts (3.07m), using a t-test with a p-value under $0.05$, see Fig.~\ref{fig:changing_friction_adaptation}. In Fig.~\ref{fig:velocity}, the curves and shaded areas are the means and standard deviations of the velocity, respectively. Snapshots of the walking gait using the meta-trained model are shown in Fig.~\ref{fig:snapshots}, more details of walking performance can be seen in \href{https://youtu.be/s9OWhxorVmc}{the video here}. 

\begin{figure}[t]
    \centering
    \includegraphics[width=\linewidth]{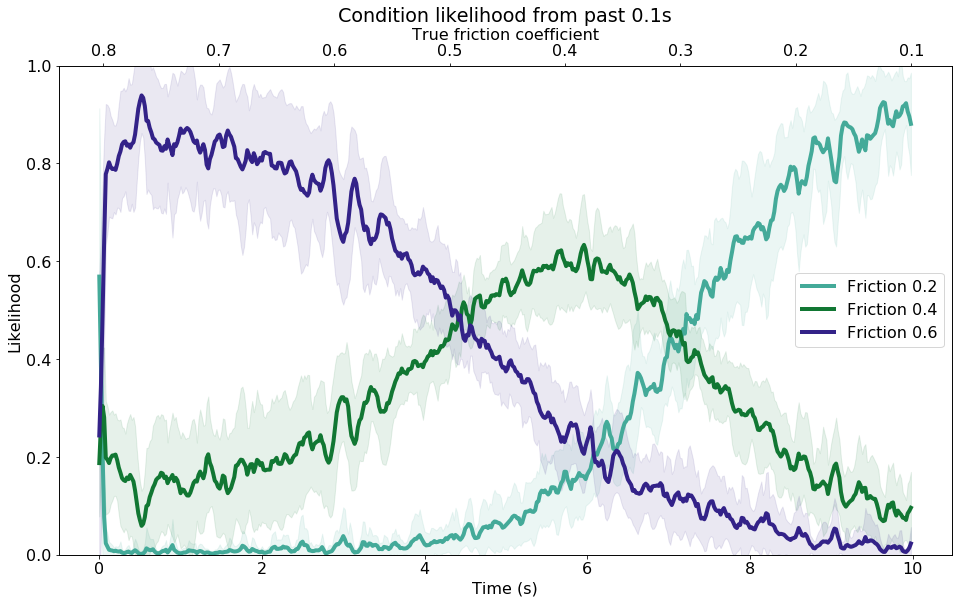}
    \vspace{-4mm}
    \caption{Likelihood of each pre-trained condition using online meta-adaptation in presence of decreasing friction.}
    \vspace{-0mm}
    \label{fig:condition_estimation_changing_friction}
\end{figure}

Additionally, Fig.~\ref{fig:condition_estimation_changing_friction} depicts the estimated condition at each time step using the past 0.1s (i.e. 5 time steps). At the beginning of the episode, when friction was higher, the model estimated a friction of 0.6 to be more likely (from 0.8 down to 0.5, i.e., 0-4.5s), then switched to a friction of 0.4 (from 0.5 down to 0.3, i.e., 4.5-7s), and finished by estimating a more likely friction of 0.2 (from 0.3 down to 0.1, i.e., 7-10s).

\subsection{Discussion}
The results validate the efficiency and effectiveness of the meta-learning method to detect the most probable current condition and adapt accordingly to different ground friction coefficients. The online adaptive walking using the proposed meta-learning outperformed the specialized experts which were specifically trained on specific frictions. This demonstrated the capability of meta-learning to incorporate knowledge from all training data.

We have also pushed the extreme test case in terms of unseen hardware failures. We specifically designed a case for meta-training where a motor of one leg of the quadruped was blocked at a fixed joint position (emulated actuator failures), and tested if meta-learning can adapt to the damage from a different leg faster than learning from scratch. Our investigation showed that the meta-adaptation is not able to adapt to such changes on a different leg, and we hypothesize that a second-order meta-learning algorithm, e.g. Model-Agnostic Meta-Learning~\cite{MAML}, may be a better solution for such an extreme case.

\section{Conclusion and future work}
\label{sec:Conclusion}
Based on the past work of model-based meta-RL~\cite{FAMLE}, we have made contributions to improve the algorithm for adaptive and robust quadruped locomotion in changing robot dynamics (motor failure and amputation) and varying environmental constraints (time-varying friction, external pushes). In physics-based simulation, we have demonstrated this method can learn quadrupedal walking without using a periodic gait signal \cite{FAMLE} or a phase vector \cite{yang2020learning}. Instead, by updating an interaction model of the robot and environment and applying the optimal control actions, a walking gait is naturally generated as the outcome of maximizing the task reward. We further validated the capability of our proposed framework in adapting to different conditions such as robot damage, changing friction, and external force disturbances. 

Future work will apply this method on the real SpotMicro robot, and identify potential issues of sim2real transfer which will be addressed by new solutions for meta-model and more effective search of model predictive control procedure. We hypothesize the current meta-learning algorithm is efficient enough for multi-task learning, which can be further studied. Also, a second-order meta-learning algorithm~\cite{MAML} can be Incorporated could potentially achieve better adaptation to novel situations.

\section{Acknowledgement}
This work has been supported by EPSRC UK Robotics and Artificial Intelligence Hub for Offshore Energy Asset Integrity Management (EP/R026173/1).


\bibliographystyle{IEEEtran}
\balance
\bibliography{IEEEabrv,main-arxiv}

\begin{thebibliography}{10}
\providecommand{\url}[1]{#1}
\csname url@samestyle\endcsname
\providecommand{\newblock}{\relax}
\providecommand{\bibinfo}[2]{#2}
\providecommand{\BIBentrySTDinterwordspacing}{\spaceskip=0pt\relax}
\providecommand{\BIBentryALTinterwordstretchfactor}{1}
\providecommand{\BIBentryALTinterwordspacing}{\spaceskip=\fontdimen2\font plus
\BIBentryALTinterwordstretchfactor\fontdimen3\font minus
  \fontdimen4\font\relax}
\providecommand{\BIBforeignlanguage}[2]{{%
\expandafter\ifx\csname l@#1\endcsname\relax
\typeout{** WARNING: IEEEtran.bst: No hyphenation pattern has been}%
\typeout{** loaded for the language `#1'. Using the pattern for}%
\typeout{** the default language instead.}%
\else
\language=\csname l@#1\endcsname
\fi
#2}}
\providecommand{\BIBdecl}{\relax}
\BIBdecl

\bibitem{fankhauser2018robust}
P.~Fankhauser \emph{et~al.}, ``Robust rough-terrain locomotion with a
  quadrupedal robot,'' in \emph{2018 IEEE International Conference on Robotics
  and Automation (ICRA)}.\hskip 1em plus 0.5em minus 0.4em\relax IEEE, 2018,
  pp. 1--8.

\bibitem{chatzinikolaidis2020}
I.~Chatzinikolaidis \emph{et~al.}, ``Contact-implicit trajectory optimization
  using an analytically solvable contact model for locomotion on variable
  ground,'' \emph{IEEE Robot. Autom. Lett.}, vol.~5, no.~4, pp. 6357--6364,
  2020.

\bibitem{rai2018bayesian}
A.~Rai \emph{et~al.}, ``Bayesian optimization using domain knowledge on the
  atrias biped,'' in \emph{2018 IEEE International Conference on Robotics and
  Automation (ICRA)}.\hskip 1em plus 0.5em minus 0.4em\relax IEEE, 2018, pp.
  1771--1778.

\bibitem{yuan2019}
K.~Yuan \emph{et~al.}, ``Bayesian optimization for whole-body control of
  high-degree-of-freedom robots through reduction of dimensionality,''
  \emph{IEEE Robot. Autom. Lett.}, vol.~4, no.~3, pp. 2268--2275, 2019.

\bibitem{dallali2012global}
H.~Dallali \emph{et~al.}, ``On global optimization of walking gaits for the
  compliant humanoid robot, coman using reinforcement learning,''
  \emph{Cybernetics and Information Technologies}, vol.~12, no.~3, pp. 39--52,
  2012.

\bibitem{yang2020learning}
C.~Yang \emph{et~al.}, ``Learning natural locomotion behaviors for humanoid
  robots using human bias,'' \emph{IEEE Robotics and Automation Letters},
  vol.~5, no.~2, pp. 2610--2617, 2020.

\bibitem{hwangbo2019learning}
J.~Hwangbo \emph{et~al.}, ``Learning agile and dynamic motor skills for legged
  robots,'' \emph{Science Robotics}, vol.~4, no.~26, 2019.

\bibitem{yang2020multi}
C.~Yang \emph{et~al.}, ``Multi-expert learning of adaptive legged locomotion,''
  \emph{Science Robotics}, vol.~5, no.~49, 2020.

\bibitem{hessel_rainbow_2017}
M.~Hessel \emph{et~al.}, ``Rainbow: Combining improvements in deep
  reinforcement learning,'' in \emph{AAAI-18}, 2018, pp. 3215--3222.

\bibitem{alphastarblog}
O.~Vinyals \emph{et~al.}, ``{AlphaStar: Mastering the Real-Time Strategy Game
  StarCraft II},'' 2019.

\bibitem{Chua}
K.~Chua \emph{et~al.}, ``Deep reinforcement learning in a handful of trials
  using probabilistic dynamics models,'' in \emph{NeurIPS 2018}, 2018, pp.
  4759--4770.

\bibitem{GrBAL}
A.~Nagabandi \emph{et~al.}, ``Learning to adapt in dynamic, real-world
  environments through meta-reinforcement learning,'' in \emph{{ICLR} 2019},
  2019.

\bibitem{FAMLE}
R.~Kaushik \emph{et~al.}, ``Fast online adaptation in robotics through
  meta-learning embeddings of simulated priors,'' 2020.

\bibitem{PPO}
J.~Schulman \emph{et~al.}, ``Proximal policy optimization algorithms,''
  \emph{CoRR}, vol. abs/1707.06347, 2017.

\bibitem{SimToReal}
J.~Tan \emph{et~al.}, ``Sim-to-real: Learning agile locomotion for quadruped
  robots,'' in \emph{Robotics: Science and Systems XIV}, 2018.

\bibitem{SAC}
T.~Haarnoja \emph{et~al.}, ``Soft actor-critic: Off-policy maximum entropy deep
  reinforcement learning with a stochastic actor,'' in \emph{{ICML} 2018},
  2018, pp. 1856--1865.

\bibitem{LearningToWalk}
T.~Haarnoja \emph{et~al.}, ``Learning to walk via deep reinforcement
  learning,'' in \emph{Robotics: Science and Systems XV}, 2019.

\bibitem{ha_learning_2020}
S.~Ha \emph{et~al.}, ``Learning to {Walk} in the {Real} {World} with {Minimal}
  {Human} {Effort},'' \emph{arXiv:2002.08550 [cs]}, Feb. 2020.

\bibitem{BayesianOptimisation}
E.~Brochu \emph{et~al.}, ``A tutorial on bayesian optimization of expensive
  cost functions, with application to active user modeling and hierarchical
  reinforcement learning,'' \emph{CoRR}, vol. abs/1012.2599, 2010.

\bibitem{noveltySearch}
J.~Lehman and K.~O. Stanley, ``Exploiting open-endedness to solve problems
  through the search for novelty,'' in \emph{Artificial Life {XI}}, 2008, pp.
  329--336.

\bibitem{MapElite}
J.-B. Mouret and J.~Clune, ``Illuminating search spaces by mapping elites,''
  \emph{arXiv preprint arXiv:1504.04909}, 2015.

\bibitem{IMGEP}
S.~Forestier \emph{et~al.}, ``Intrinsically motivated goal exploration
  processes with automatic curriculum learning,'' \emph{CoRR}, vol.
  abs/1708.02190, 2017.

\bibitem{levineLinearModel}
S.~Levine \emph{et~al.}, ``End-to-end training of deep visuomotor policies,''
  \emph{J. Mach. Learn. Res.}, vol.~17, pp. 39:1--39:40, 2016.

\bibitem{WorldModel}
D.~Ha and J.~Schmidhuber, ``Recurrent world models facilitate policy
  evolution,'' in \emph{NeurIPS 2018}, 2018, pp. 2455--2467.

\bibitem{Pilco}
M.~Deisenroth and C.~Rasmussen, ``Pilco: A model-based and data-efficient
  approach to policy search,'' in \emph{ICML 2011}, 2011, pp. 465--472.

\bibitem{Multi-dex}
R.~Kaushik \emph{et~al.}, ``Multi-objective model-based policy search for
  data-efficient learning with sparse rewards,'' in \emph{CoRL 2018}, vol.~87,
  2018, pp. 839--855.

\bibitem{Gal2015DropoutB}
Y.~Gal and Z.~Ghahramani, ``Dropout as a bayesian approximation: Representing
  model uncertainty in deep learning,'' in \emph{{ICML} 2016}, vol.~48, 2016,
  pp. 1050--1059.

\bibitem{MPPI}
G.~Williams \emph{et~al.}, ``Model predictive path integral control using
  covariance variable importance sampling,'' \emph{CoRR}, vol. abs/1509.01149,
  2015.

\bibitem{CEM}
R.~Y. Rubinstein and D.~P. Kroese, \emph{The {Cross}-{Entropy} {Method}: {A}
  {Unified} {Approach} to {Combinatorial} {Optimization}, {Monte}-{Carlo}
  {Simulation} and {Machine} {Learning}}, 2004.

\bibitem{learningToLearn}
K.~Li and J.~Malik, ``Learning to optimize,'' in \emph{{ICLR}}, 2017.

\bibitem{MAML}
C.~Finn \emph{et~al.}, ``Model-agnostic meta-learning for fast adaptation of
  deep networks,'' in \emph{International Conference on Machine Learning,
  {ICML}}, vol.~70.\hskip 1em plus 0.5em minus 0.4em\relax {PMLR}, 2017, pp.
  1126--1135.

\bibitem{ES-ML}
X.~Song \emph{et~al.}, ``Rapidly adaptable legged robots via evolutionary
  meta-learning,'' \emph{arXiv preprint arXiv:2003.01239}, 2020.

\bibitem{ml-gp}
S.~S{\ae}mundsson \emph{et~al.}, ``Meta reinforcement learning with latent
  variable gaussian processes,'' \emph{arXiv preprint arXiv:1803.07551}, 2018.

\bibitem{noRML}
Y.~Yang \emph{et~al.}, ``Norml: No-reward meta learning,'' \emph{CoRR}, 2019.

\bibitem{M3PO}
T.~Hiraoka \emph{et~al.}, ``Meta-model-based meta-policy optimization,''
  \emph{arXiv preprint arXiv:2006.02608}, 2020.

\bibitem{adMRL}
Z.~Lin \emph{et~al.}, ``Model-based adversarial meta-reinforcement learning,''
  \emph{arXiv preprint arXiv:2006.08875}, 2020.

\bibitem{context-variable}
R.~Mendonca \emph{et~al.}, ``Meta-reinforcement learning robust to
  distributional shift via model identification and experience relabeling,''
  \emph{arXiv preprint arXiv:2006.07178}, 2020.

\bibitem{bias-transformation}
C.~Finn \emph{et~al.}, ``One-shot visual imitation learning via
  meta-learning,'' \emph{CoRR}, 2017.

\bibitem{adam}
D.~P. Kingma and J.~Ba, ``Adam: {A} method for stochastic optimization,'' in
  \emph{{ICLR} 2015,}, 2015.

\bibitem{reptile}
A.~Nichol \emph{et~al.}, ``On first-order meta-learning algorithms,''
  \emph{CoRR}, 2018.

\bibitem{kim_2019}
\BIBentryALTinterwordspacing
D.-Y. Kim, ``Spotmicro - robot dog by kdy0523,'' Feb 2019. [Online]. Available:
  \url{https://www.thingiverse.com/thing:3445283}
\BIBentrySTDinterwordspacing

\end{thebibliography}

\end{document}